\documentclass[letterpaper, 10 pt, conference]{ieeeconf} 
\IEEEoverridecommandlockouts
\usepackage{cite}
\usepackage{lmodern}
\usepackage{graphicx}
\usepackage{times}
\usepackage{ragged2e}
\usepackage{amsmath,amssymb, amsfonts}
\usepackage{tabularx}
\usepackage{hyperref}
\usepackage{cleveref}
\usepackage[ruled,vlined]{algorithm2e}
\usepackage{siunitx}
\usepackage{dblfloatfix}
\usepackage{authblk}
\usepackage[table]{xcolor}
\usepackage[T1]{fontenc}
\usepackage{siunitx}
\usepackage{booktabs}

\begin{document}
\noindent

\title{\LARGE \bf
A Robust Hybrid Observer for Side-slip Angle Estimation  
}

\author{
Agapius Bou Ghosn$^{1}$,
Marcus Nolte$^{2}$,
Philip Polack$^{1}$,
and Arnaud de La Fortelle$^{1,3}$
\thanks{$^{1}$ Center for Robotics, Mines Paris, PSL University, 75006 Paris, France {\tt [agapius.bou\textunderscore ghosn, philip.polack, arnaud.de\textunderscore la\textunderscore fortelle]@minesparis.psl.eu}}
\thanks{$^{2}$ Institute for Control Engineering, TU Braunschweig, 38106 Braunschweig, Germany {\tt nolte@ifr.ing.tu-bs.de}}
\thanks{$^{3}$ Heex Technologies, Paris, France}
}
\maketitle

\thispagestyle{empty}
\pagestyle{empty}

\begin{abstract}
For autonomous driving or advanced driving assistance, it is key to monitor the vehicle dynamics behavior. Accurate models of this behavior include acceleration, but also the side-slip angle, that eventually results from the complex interaction between the tires and the road. Though it is an essential quantity (e.g. for stability assessment), as opposed to accelerations, it is not measurable through conventional off-the-shelf sensors. Therefore, accurate side-slip angle observers are necessary for the proper planning and control of vehicles. In this paper, we introduce a novel approach that combines model-based side-slip angle estimation with neural networks. We apply our approach to real vehicle data. We prove that the proposed method is able to outperform state-of-the-art methods for normal driving maneuvers, and for near-limits maneuvers where providing accurate estimations becomes challenging. 
\end{abstract}

\section{Introduction}
The side-slip angle of a vehicle is defined as the angle between the velocity vector and its longitudinal axis; it is a fundamental quantity to assess vehicle stability and by that an important indicator for critical driving situations \cite{kiencke_observation_1997}. Moreover, it is part of the state in many non-linear models of vehicle dynamics. However, the side-slip angle is only measurable with expensive sensors, either based on optical flow directly over ground or highly accurate dual-antenna GNNS solutions. For this reason, the side-slip angle is a typical application for state observers in the literature.
%The inaccessibility of the side-slip angle through conventional vehicle sensors made its knowledge in the literature linked to state observers. 

A classic implementation of state observers is model-based, relying on a physical model describing the vehicle dynamics' behavior along with the available measurements (e.g. from inertial measurement units and/or cameras \cite{dickmanns1992}).
Other approaches are based on neural networks (or learning-based), using training data to parameterize a highly non-linear black-box model that provides a direct input-output mapping from measurement data to the estimated dynamic states. 
In the first case, the accuracy of the observer is related to the quality of the vehicle model and the observer's algorithm.
In the second case the accuracy is related to the quality of the training data and the network's architecture. 

On the one hand, model-based approaches explicitly represent physical relations and hence provide more insight and exaplainability.
However, model accuracy is always limited.
Particularly when highly non-linear tire dynamics start dominating the behavior of the overall vehicle dynamics, the estimation accuracy of model-based approaches degrades either due to non-modeled effects, or due to inevitable parameter uncertainty.
Neural-networks on the other hand are typically designed for performing accurate non-linear regression, making them especially suitable for non-linear estimation tasks -- at the cost of losing physical explainability.

For this reason, hybrid approaches that combine physical models with neural networks are becoming more and more popular (cf. \cite{graber_hybrid_2019}) as they remain physically explainable, while being able to adapt to parameter uncertainty or modeling errors.

\begin{figure}
    \centering
    \includegraphics[width=.8\columnwidth]{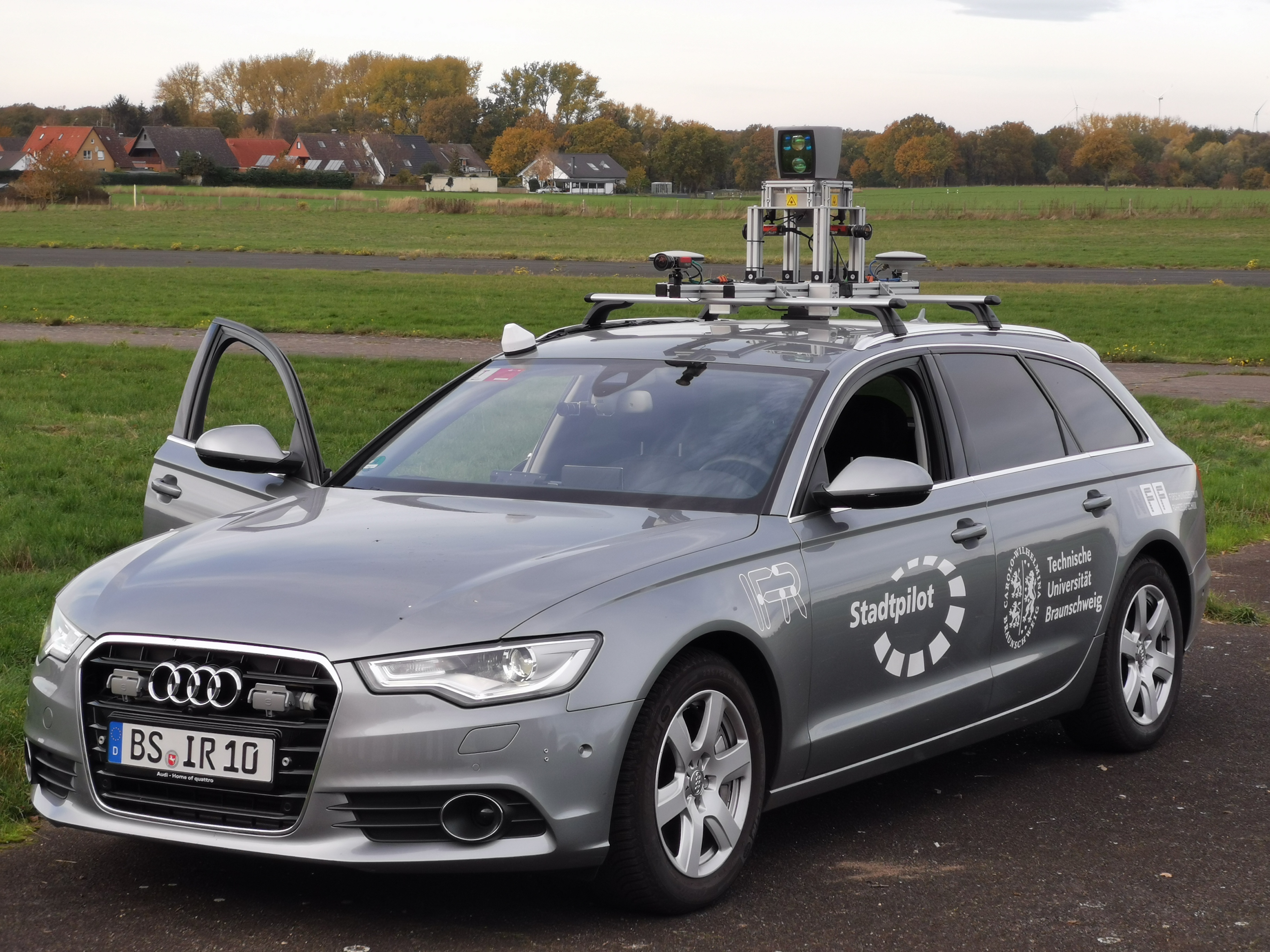}
    \caption{Stadtpilot vehicle used for testing our approach.}
    \label{vehicle.fig}
\end{figure}

This paper introduces a hybrid observer that combines kinematics and neural networks to provide reliable side-slip angle estimations in both normal and harsh driving maneuvers. The introduced observer is tested on the Stadtpilot vehicle shown in Fig. \ref{vehicle.fig}. 

The literature detailed in the next section considers state-of-the-art deterministic and learned approaches to estimate the side-slip angle of the vehicle.
 
In summary, we will prove that the proposed method is able to estimate the vehicle's side-slip angle in both low and high acceleration maneuvers using only in-car sensor measurements, outperforming state-of-the-art approaches.
For comparability, we focus on approaches that rely on inertial and GNSS sensors.

In the following, the state-of-the-art observers are presented in Section \ref{soa.sec}, the system setup is described in Section \ref{systemSetup.sec}, the used training and testing data sets are described in Section \ref{dataset.sec}, the proposed method is presented in Section \ref{approach.sec}, the results are presented and discussed in Section \ref{results.sec}; the article is concluded in Section \ref{conclusion.sec}.

\section{Related Work}\label{soa.sec}
Observers presented in the literature will be detailed in this Section. As we distinguish between classical (or deterministic) observers and learning-based observers, the presented literature will be split into these two categories in Subsections \ref{deterministicLit.ssec} and \ref{learnedLit.ssec} respectively. Subsection \ref{learnedLit.ssec} also includes hybrid approaches. After presenting the current state-of-the-art approaches, we will conclude by choosing the observers to which our method will be compared. 

\subsection{Classical observers}\label{deterministicLit.ssec}
Classical observers are extensively used in the literature to estimate the states and parameters of a vehicle. They rely on a model used to describe the state evolution of the vehicle. The complexity of the model used differs between applications and can vary between using simple kinematic models as in \cite{panzieri_outdoor_2002} to four wheel dynamic models with complex tire models (e.g. Pacejka tire model) as in \cite{wenzel_dual_2006}. The choice of the model used for estimation determines the number of quantities to be known to estimate the required state, as well as the governing assumptions. The used model determines the domain of validity of the observer. 

Different types of model based observers could be used as the Luenberger observer, which is a linear observer, used in \cite{kiencke_observation_1997} and \cite{cherouat_vehicle_2005}, for the estimation of the vehicle velocity, side slip angle, and yaw rate; in these works, a dynamic bicycle model and a linear tire model are used, but inadequacies occur when the tire model is no longer valid.
Nonlinear observers dealing with nonlinear models, as in \cite{lee_slip_2013}, \cite{wang_estimation_2010}, \cite{song_pneumatic_2014} are used to estimate the side slip angle; in these works, the model used is either a dynamic bicycle model or a four wheel dynamic model with nonlinear tire models; the used models ensure a wider representation of the vehicle's maneuvers, resulting in fewer estimation errors in nonlinear cases.
While Luenberger and related observers solve initial value problems, besides their restriction to linear plant models, they cannot account for model uncertainty (or process noise) or measurement noise.  

The Kalman filter solves some of the limitations of the Luenberger observer. In its basic implementation, it can only be applied to linear systems. In contrast to the Luenberger observer, it has been specifically designed to account for model uncertainty and (Gaussian) measurement noise. 
%The Kalman filter procedure is split into a prediction step that predicts the current state and uncertainty based on the previous state, applied inputs and covariances and an update step where the predicted state is corrected by the current measurement. 
The Extended Kalman Filter (EKF) is an extension to the Kalman filter to be applied to nonlinear systems. It linearizes at each step around the current estimate. The EKF has been used with a dynamic bicycle prediction model or a four wheel dynamic prediction model for the estimation of the vehicle's slip angle as in \cite {dickmanns1992}, \cite{van_aalst_adaptive_2018}, \cite{reina_vehicle_2019}  with different tire models.

\subsection{Learned observers}\label{learnedLit.ssec}
Learned observers have recently been presented for data-based vehicle state estimation. They are implemented either as hybrid observers, combining deterministic equations and neural networks, or as fully learned observers involving only neural networks. 

Hybrid approaches can be split into two types. On the one hand, several approaches combine a model-based filter with neural networks, such as the KalmanNet \cite{revach_kalmannet_2021}.
KalmanNet integrates recurrent neural networks in a Kalman-filter-like predictor-corrector structure; it is employed in \cite{escoriza_data-driven_2021} to estimate vehicle velocities in x- and y direction. \cite{noauthor_vehicle_2021} combine a sliding mode observer and a neural network to estimate the vehicle's velocities. 
The second type of approaches applies  vehicle dynamics equations to support the neural-networks' estimations by providing additional inputs into the network. \cite{graber_hybrid_2019} e.g. calculate the side-slip angle rate based on a single-track model and feed it as an input to the neural network along with the measurements to enhance  the network's side-slip angle estimation. 

Fully learned approaches implement Long Short-Term Memory (LSTM) or Gated Recurrent Units (GRU) networks, as in  \cite{zhang_reliable_2021} and \cite{srinivasan_end--end_2020} where a vehicle's velocities are predicted from sensor measurements, or\cite{bonfitto_combined_2020} where the vehicle's side slip angle is predicted from sensor measurements. 

As we are interested in estimating the vehicle's side-slip angle, we will compare our observer to the EKF based approach introduced in \cite{reina_vehicle_2019} and to the hybrid approach introduced in \cite{graber_hybrid_2019}, knowing that both approaches showed accurate estimations in their testing. The work done in \cite{bonfitto_combined_2020} proposes a classification approach to identify the road conditions which is out of the scope of this paper. 

It should be noted that a fully learned approach, with no physical model input, was considered in \cite{graber_hybrid_2019} and was proven to have higher errors than the hybrid approach that we compare to: this is why no comparison with fully learned methods is proposed in our paper.  

For the comparison, we implemented the approaches presented in \cite{reina_vehicle_2019} and \cite{graber_hybrid_2019}.
We trained the hybrid approach (\cite{graber_hybrid_2019}) on our own training data set and fed all implementations with the same data from our testing set.

\section{System Setup}\label{systemSetup.sec}
As we present a side-slip angle observer for autonomous vehicles, we applied this approach on a real vehicle: the Stadtpilot vehicle (AUDI A6 Avant C7) shown in Fig. \ref{vehicle.fig}. The characteristics of the vehicle are shown in Table \ref{vehicleChs.tab}.

\begin{table}
\vspace{.2cm}
\centering
%  \resizebox{\columnwidth}{!}{
    \caption{Parameters of the vehicle used for data collection (CoG: center of gravity)}
    \begin{tabular}{cp{.5\columnwidth}c}
      \toprule
      \textbf{Parameter} & \textbf{Description} & \textbf{Value} \\
      \midrule
      $M$ & Mass of the vehicle & \SI{1578}{\kilo\gram}\\
      $l_\mathrm{f}$ & length from CoG to the front axle & \SI{1.134}{\meter} \\
      $l_\mathrm{r}$ & length from CoG to the rear axle  & \SI{1.578}{\meter} \\
      $b$ & Track width & \SI{1.513}{\meter} \\
      $I_z$ & Moment of inertia around the $z$-axis & \SI{2924}{\kilo\gram\square\meter}\\
      \bottomrule
    \end{tabular}
%}
\label{vehicleChs.tab}
\end{table}

\begin{figure}
    \vspace{.2cm}
    \centering
    \includegraphics[width=.85\columnwidth]{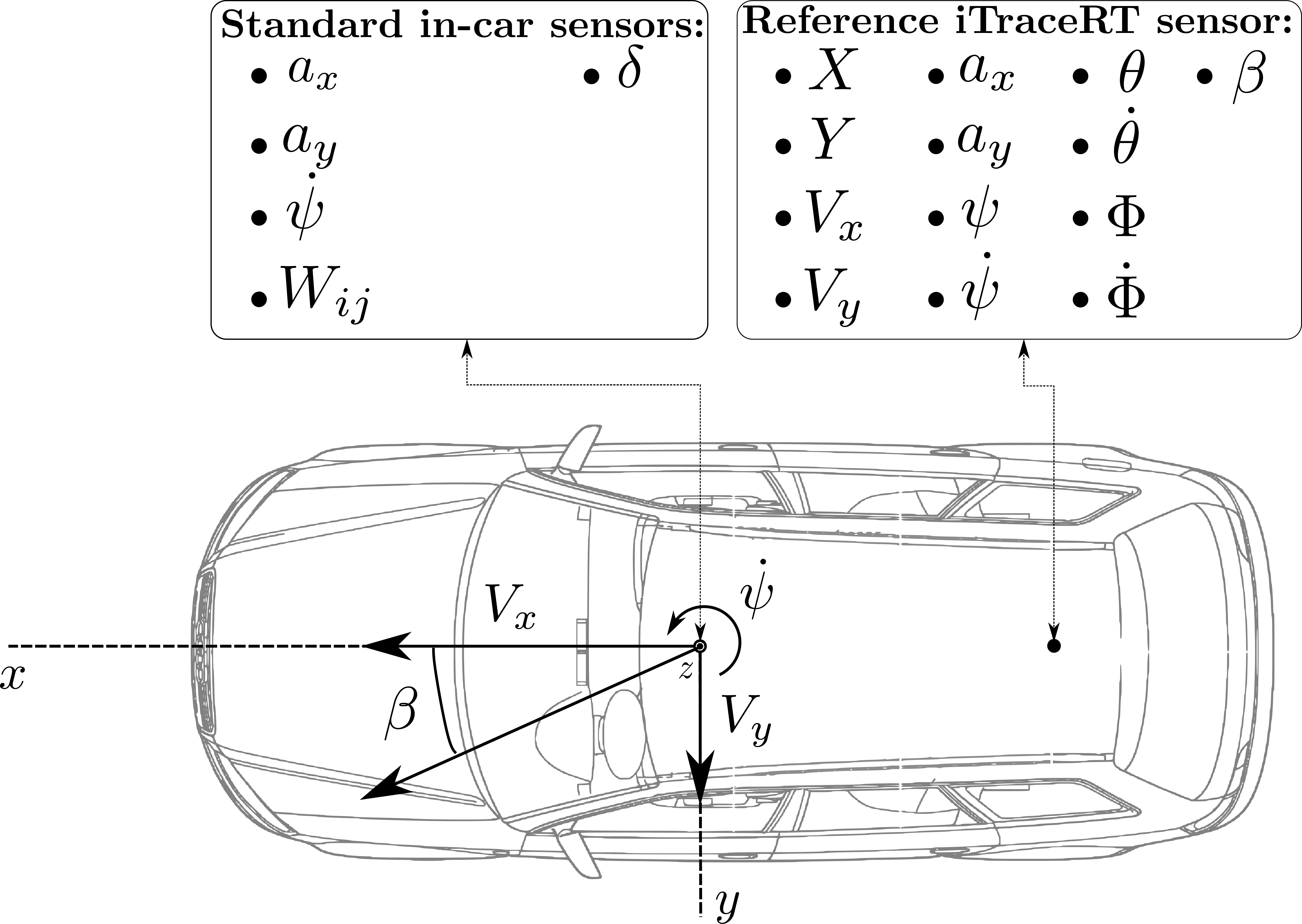}
    \caption{Top view of the used vehicle showing the available in-car and reference sensor measurements. The side-slip angle $\beta$ shown at the center of gravity of the vehicle will be estimated. The variables are described in Section \ref{systemSetup.sec}.}
    \label{topView.fig}
\end{figure}

The experimental platform is the same that we used for collecting data for a non-hybrid observer structure in \cite{bou_ghosn_robust_2023}. For the sake of readability, we provide a very similar description: The vehicle is equipped with a conventional inertial measurement unit (IMU), the Audi Sensor Array (SARA), which measures the longitudinal and lateral accelerations ($a_x$, $a_y$), the yaw rate ($\dot\psi$), and the wheel speeds ($W_{ij}$). Steering angle measurements ($\delta$) are also available. A reference dual-antenna INS/GNSS \href{https://www.imar-navigation.de/en/products/by-product-names/item/itracert-f200-itracert-f400-itracert-mvt}{iTraceRT F400} sensor provides highly accurate measurement of the position (Easting -- $X$ , Northing -- $Y$) in UTM-coordinates, the longitudinal and lateral velocities ($V_x$, $V_y$), the longitudinal and lateral accelerations ($a_x$, $a_y$), the yaw ($\psi$), pitch ($\theta$), and roll ($\Phi$) angles and rates ($\dot\psi$, $\dot\theta$, $\dot\Phi$), as well as the side-slip angle ($\beta$). The top view of the used vehicle is presented in Fig. \ref{topView.fig}; it presents the sensors mounted on the vehicle with the measurements they provide in addition to the side-slip angle to be estimated at the center of gravity. The Audi SARA IMU provides measurements at a \SI{50}{Hz} frequency while the iTraceRT sensor provides measurements at \SI{100}{Hz}; The measurements provided by the iTraceRT sensor are down sampled to \SI{50}{Hz} for synchronization purposes. 
The developed observer takes its input data from the in-car sensors and compares its outputs to the ground truth measurements provided by the iTraceRT sensor. 
Data collection by the described sensors is presented next.

\section{Data Set}\label{dataset.sec}
Our data set is split into two categories: The first one is composed of normal driving through out the city of Braunschweig, Germany in normal weather conditions; the second one is composed of harsh maneuvers executed on a special test track near Peine, Germany in varying weather conditions. We collected 1.03 million data samples in 5.7 hours of driving. 

\begin{figure}
\vspace{.2cm} 
    \centering
    \includegraphics[width=.85\columnwidth]{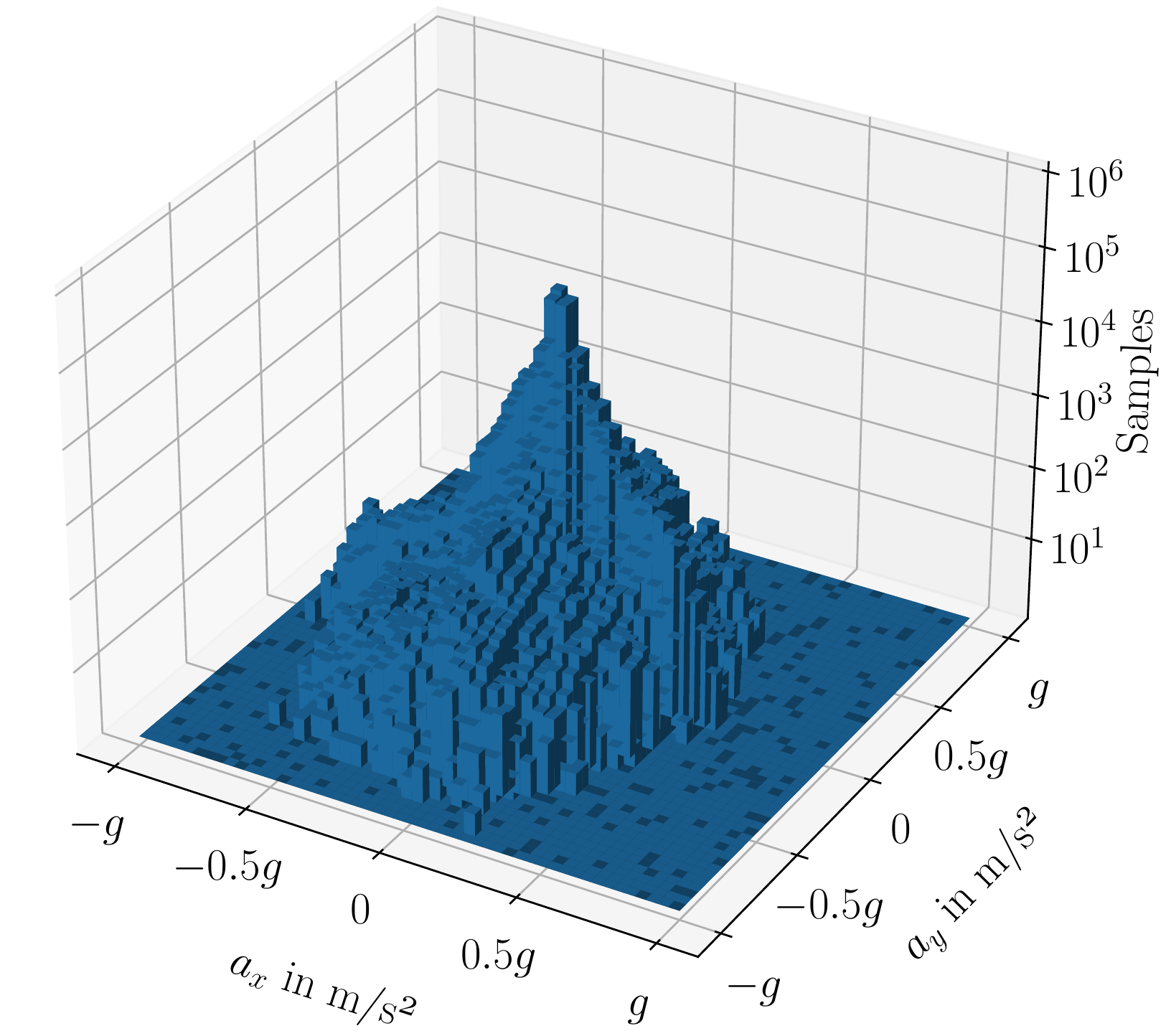}
    \caption{Distribution of the collected data on the friction circle (number of samples shown on a logarithmic axis): Many of the maneuvers are in low acceleration range, but samples are also present at higher accelerations. $g$ is the gravitational acceleration expressed in $\si{\meter / \second^{2}}$.}
    \label{frictionCircle.fig}
\end{figure}

\begin{figure}
    \centering
    \includegraphics[width=.9\columnwidth]{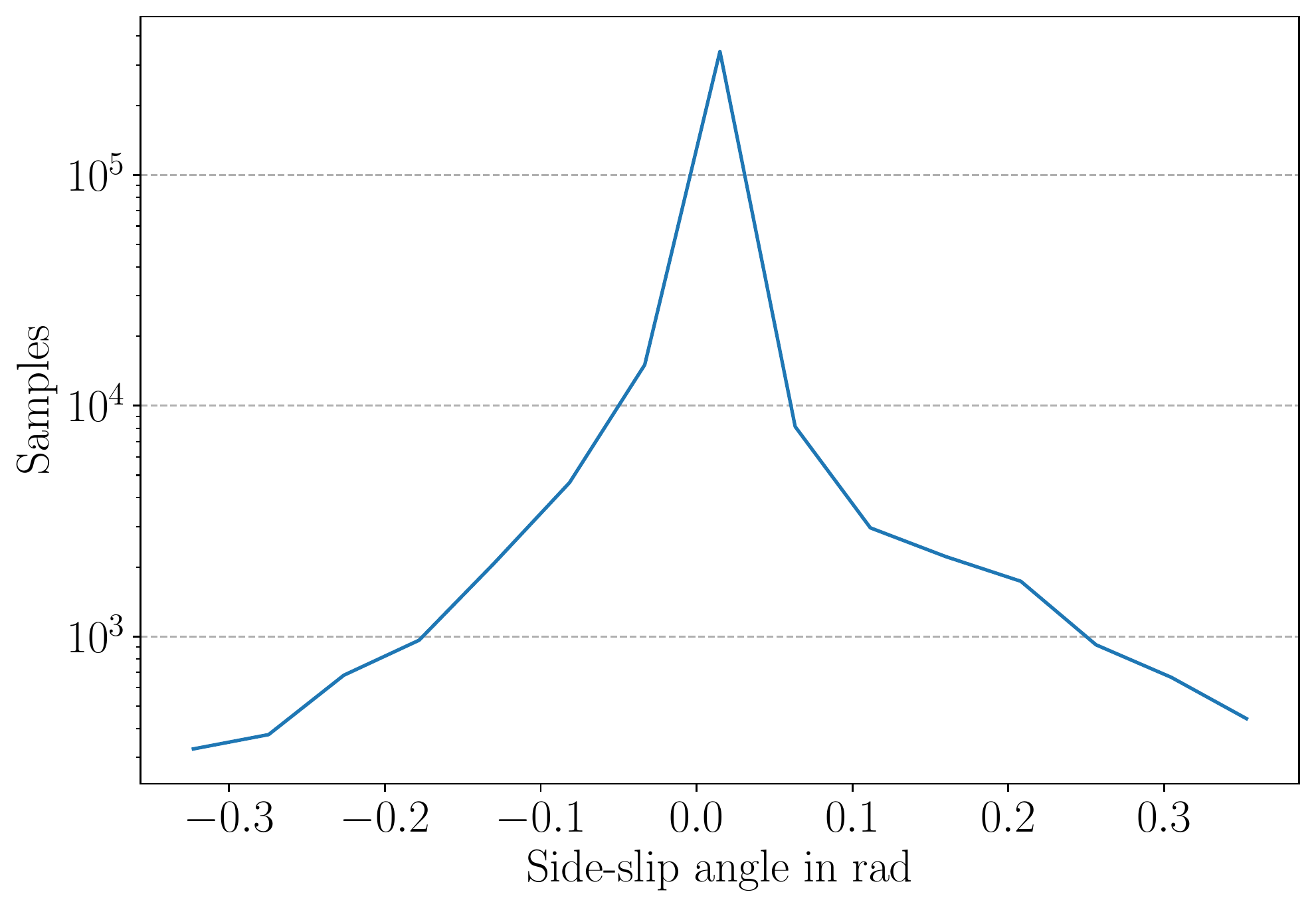}
    \caption{Distribution of the side-slip angles over the collected data set. Small side-slip angle data is easier to collect which explains the distribution being biased towards smaller angles.}
    \label{samplesAngle.fig}
\end{figure}

Fig. \ref{frictionCircle.fig} presents accumulated acceleration measurements in the friction circle, to give an impression of the vehicle dynamics that we were able to achieve during data collection. The plot shows that most of the effected maneuvers are within the low acceleration range which corresponds to the normal city driving and other maneuvers at low speed while other samples are in the higher acceleration range (reaching $a=1g$), those correspond to the harsh maneuvers on the test track. The distribution is biased towards lower accelerations due to the ease of collecting data at lower accelerations and the difficulty of collecting data at the limits of handling; this is also reflected in the distribution of the side-slip angles over the collected data set shown in Fig. \ref{samplesAngle.fig}. It can be seen that the dataset contains a near-symmetrical distribution of side-slip angles in a range von approximately $\pm \SI{18}{\degree}$. As the sample count is comparatively low for high side-slip angles, we will consider the accuracy of our estimation specifically in high-dynamic maneuvers.

An 80\%, 20\% split separates the training data set from the testing data set; the separation is performed such that both data sets include low acceleration and high acceleration data. 
Having defined the data sets, the proposed approach is presented next.

\section{Proposed Approach}\label{approach.sec}
The goal of the developed observer is to estimate the side-slip angle of the vehicle given the available measurements. We propose a hybrid architecture that relies on a kinematic bicycle model and neural networks to provide accurate estimations. The kinematic bicycle model is presented next, followed by the observer's proposed architecture.

\subsection{The kinematic bicycle model}
\begin{figure}
    \centering
    \includegraphics[width=.8\columnwidth]{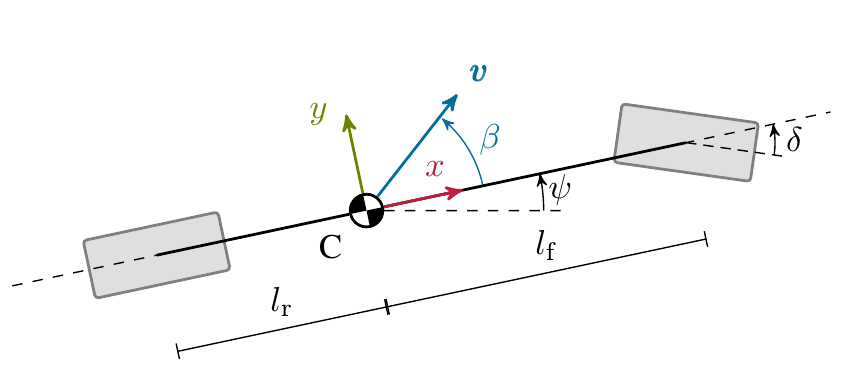}
    \caption{The kinematic bicycle model. $C$ is the center of gravity; $l_f$ and $l_r$ are the distances between the front and rear axles and the center of gravity; $v$ is the velocity vector at the center of gravity; $\delta$ is the steering angle.}
    \label{kbm.fig}
\end{figure}

The kinematic bicycle, or single-track, model shown in Fig. \ref{kbm.fig} is a simplified vehicle model valid for low speed applications. It is governed by several assumptions: first, the four-wheel vehicle is simplified into a bicycle model: front wheels are considered as a single steerable wheel, rear wheels as a single non-steerable wheel; second, the pitch and roll dynamics are neglected; third, the aerodynamic forces and the effect of the longitudinal wheel forces on the lateral dynamics are neglected.
The side-slip angle $\beta$ of the kinematic bicycle model is then defined by:
\begin{equation}
    \beta_{\text{kinematic}} = \arctan\left(\frac{l_r \tan\delta}{l_f+l_r}\right)
\end{equation}
% To examine the performance of the side-slip angle provided by the linear single-track model, we plot it along the reference iTraceRT side-slip angle (Fig. \ref{kinematicPlot.fig}). This shows that the kinematic side-slip angle is able to follow the shape of the reference side-slip angle, however offsets are clearly visible. This behavior is expected as the kinematic bicycle model is a simple model only valid for low speed scenarios. 

The kinematic side-slip angle will be fed to the proposed architecture. As the error of the kinematic side-slip angle is prone to increase with harsh maneuvers \cite{polack_kinematic_2017}, the neural network is anticipated to adapt and make use of the provided input as much as possible to provide accurate estimations.

% \begin{figure}
%     \centering
%     \includegraphics[width=\columnwidth]{Figures/Kinematic-1.pdf}
%     \caption{Comparison between the behavior of the kinematic side-slip angle and the reference iTraceRT sensor: the kinematic angle is able to catch the shape of the actual angle but shows offsets.}
%     \label{kinematicPlot.fig}
% \end{figure}

\subsection{Proposed hybrid observer architecture}

\begin{figure}
    \centering
    \vspace{.2cm}
    \includegraphics[width=.85\columnwidth]{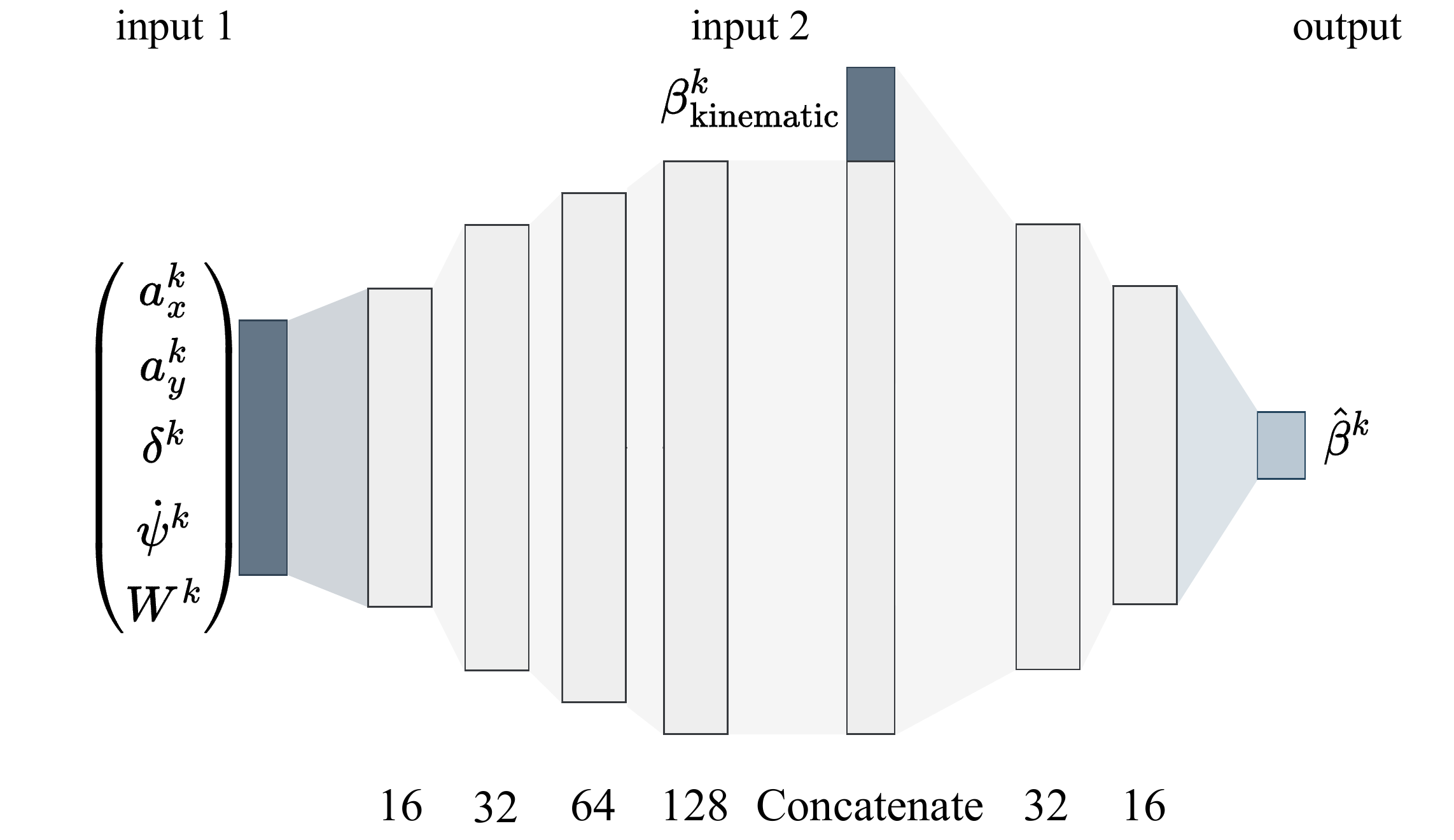}
    \caption{Proposed Architecture.}
    \label{arch.fig}
\end{figure}

The proposed hybrid observer is a multi-layer perceptron (MLP). The inputs of the observer are the vehicle's measurements at a first stage, and the kinematic side-slip angle at a second stage, both at the current time step $k$. The output of the observer is the estimated side-slip angle at the current time step $k$. The proposed architecture is shown in Fig. \ref{arch.fig}. 

The model includes in its first stage four layers with 16, 32, 64 and 128 neurons respectively; the output of the fourth layer is concatenated with the kinematic side-slip angle then fed, in a second stage, to two layers of 32 and 16 neurons respectively; the output layer follows. All the activation functions are hyperbolic tangent (tanh) except for a linear activation function at the output layer. A grid search was performed to choose the sizes of the layers. The model weights are initialized using the Xavier initialization. The loss function used is defined by:

\begin{equation}
    L = L_{\beta} + L_{\text{reg}}
\end{equation}
with $L_{\beta} = \textsc{MSE}(\hat\beta, \beta_{\text{ref}})$ and $L_{\text{reg}}$ being an L2 regularization for the network with a regularization rate of $10^{-5}$.

The network is implemented using PyTorch and is trained using an Adam optimizer on a Nvidia Geforce GTX 1650 Ti for 30 epochs with a batch size of 64. 

Note that feeding the kinematic side-slip angle at the first stage (input 1), showed higher testing errors (up to 18\%).

The trained network is tested and compared to state-of-the-art approaches next.

\section{Results}\label{results.sec}
In this section, the proposed architecture is tested on the previously defined testing set. The results are compared to the state-of-the-art approaches considered in Section \ref{soa.sec}. 

In what follows, the used metric is the mean absolute error (MAE). The proposed approach and the considered state-of-the-art approaches will be first evaluated on the whole testing set, then separate scenarios will be considered for a more detailed assessment of their performance. 

\subsection{Overall performance}
We start our evaluation by assessing the performance of the observers on the whole testing set. Thus, the MAE are calculated and shown in Table \ref{MAE1.tab}. The table shows that our approach has the lowest errors; the hybrid approach \cite{graber_hybrid_2019} is better than the EKF approach \cite{reina_vehicle_2019} but presents almost 1.4 times higher errors than our approach. 

To further examine our method, we split the following evaluation to two types of scenarios: a normal driving maneuver in which the physical models should be valid (based on analysis done in \cite{kiencke_observation_1997},\cite{polack_kinematic_2017}) and a harsh maneuver in which the physical models are expected to lose accuracy. 

\begin{table}
\centering
\setlength\tabcolsep{16 pt}
\caption{Mean absolute error (MAE) for the different observers calculated for the whole testing set. The errors of our approach are the lowest among the state-of-the-art observers. \textbf{Legend:} DBM: EKF based on the dynamic bicycle model; GRU: Hybrid GRU-based observer.\label{MAE1.tab}}
 % \resizebox{\columnwidth}{!}{
    \begin{tabular}{|c|c|c|c|c|c|}
      \hline
      State & DBM \cite{reina_vehicle_2019} &  GRU \cite{graber_hybrid_2019}  & Ours\\
      \hline
      $\beta$ \SI{}{(mrad)} & 4.27 & 3.25 & \textbf{2.37} \\
      \hline
      \end{tabular}
      % }
\end{table}

\subsection{Normal driving maneuver}
Next, we consider a "normal" driving maneuver that does not involve high excitation and which lateral accelerations are less than $a_y=0.5g$ (for model validity reasons explained in \cite{kiencke_observation_1997} and \cite{polack_kinematic_2017}). The considered maneuver has a maximum lateral acceleration $a_y^{\text{max}}=0.35g$; it is collected during the city driving phase in Braunschweig, Germany. The MAE of the different observers for the considered trajectory are shown in Table \ref{MAE2.tab}: they show that our approach beats the references by a factor of about 1.3 to 2.2 on average. 

\begin{figure}
    \centering
    \includegraphics[width=.9\columnwidth]{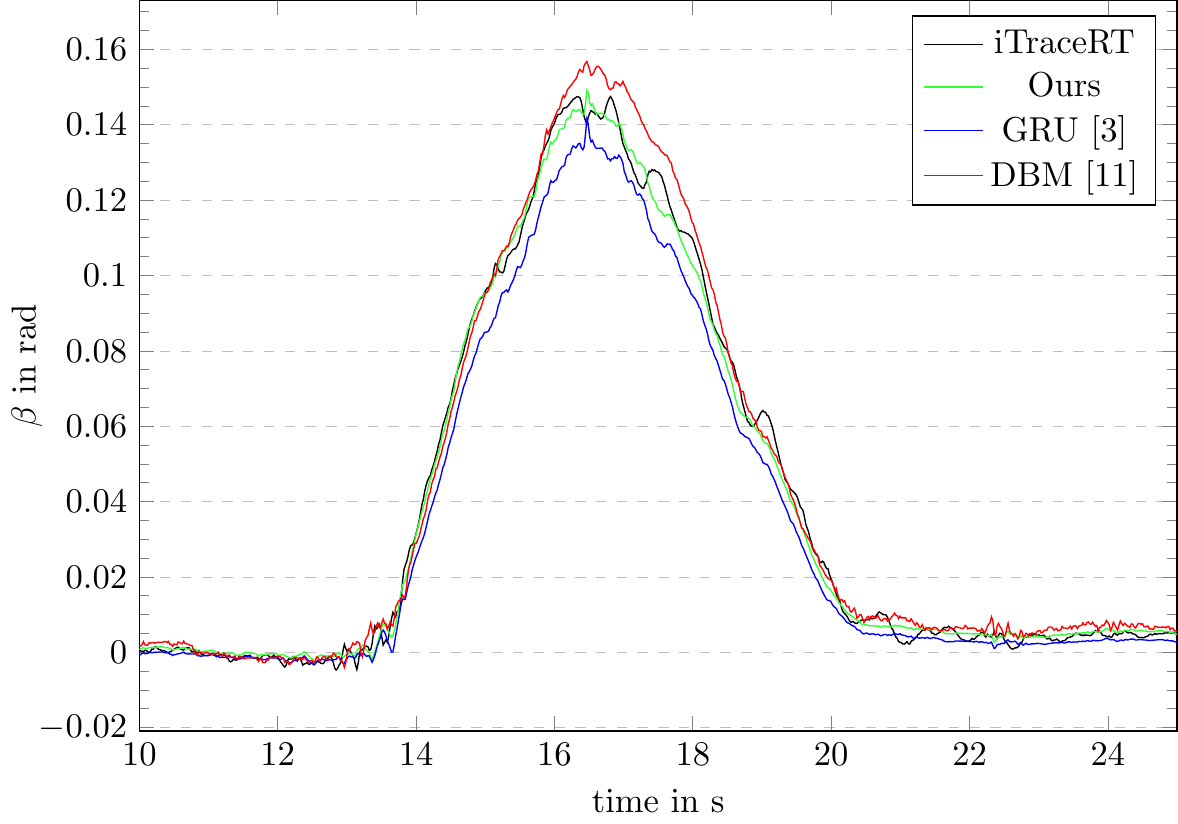}
    \caption{Performance of the different observers for the highest side-slip angle rise of the normal driving maneuver. The proposed approach is able to stick closer to the reference iTraceRT sensor than the remaining approaches.}
    \label{plot_13.fig}
\end{figure}

To visualize the behavior of the observers, we plot their estimations for the time span where the peak side-slip angle is reached: the plot is shown in Fig. \ref{plot_13.fig}. The plot shows that the proposed approach is able to observe the side-slip angle with the lowest errors. The EKF based on the dynamic bicycle model \cite{reina_vehicle_2019} shows close behavior to the proposed approach for the increasing part of the plot but delivers higher errors for the peak and the decreasing part of the plot. The GRU-based model \cite{graber_hybrid_2019} is undershooting through the whole plot. 

In the next section, we evaluate our approach in dynamic driving maneuvers.

\begin{table}
 \vspace{.2cm}
\centering
\setlength\tabcolsep{16 pt}
\caption{MAE for the different observers calculated for the normal driving maneuver ($a_y^{\text{max}} = 0.35g $). The errors of our approach are the lowest among the state-of-the-art observers. Check Table \ref{MAE1.tab} for the legend.\label{MAE2.tab}}
 % \resizebox{\columnwidth}{!}{
    \begin{tabular}{|c|c|c|c|c|c|}
      \hline
      State & DBM \cite{reina_vehicle_2019} &  GRU \cite{graber_hybrid_2019}  & Ours\\
      \hline
      $\beta$ \SI{}{(mrad)} & 3.98 & 2.35 & \textbf{1.76} \\
      \hline
      \end{tabular}
      % }

\end{table}

\subsection{Dynamic driving maneuver}
We define a dynamic driving maneuver as one that includes high vehicle excitations with lateral accelerations reaching values higher than $0.5g$. This maneuver was executed on a special testing track as described in Section \ref{dataset.sec} with a maximum lateral acceleration of $a_y^{\text{max}}=0.85g$. The MAE for the different approaches is shown in Table \ref{MAE3.tab}. All the observers present errors higher than the previous case: this is caused by the harshness of the maneuver. Our method is, however, still able to deliver the lowest errors while the EKF approach \cite{reina_vehicle_2019} delivers the highest errors. 

\begin{figure}
    \centering
    \includegraphics[width=.9\columnwidth]{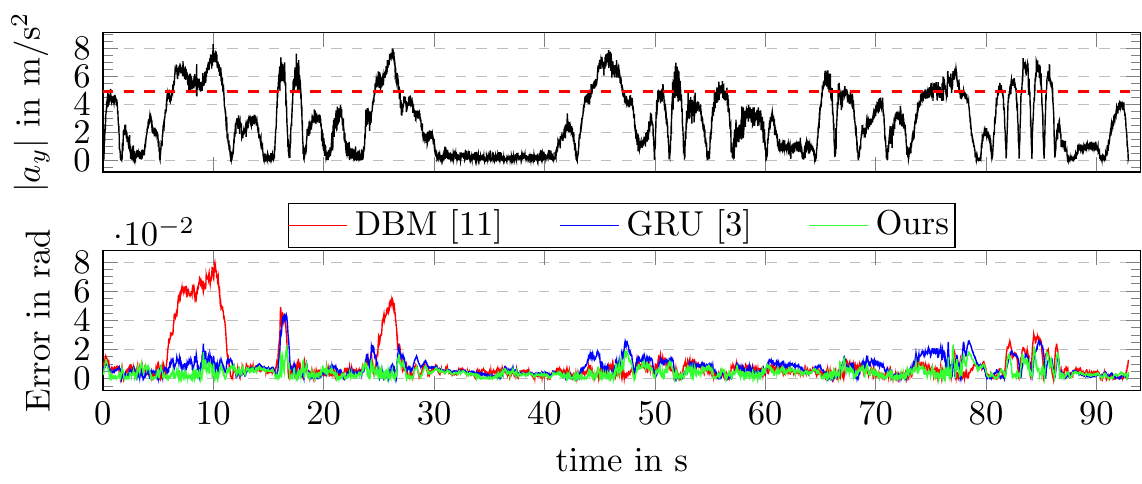}
    \caption{Errors of the different observers compared to the evolution of the lateral accelerations along the harsh maneuver. The proposed approach shows the lowest errors while the EKF based on the dynamic bicycle model shows the highest errors.}
    \label{errors_81.fig}
\end{figure}

\begin{figure}
 \vspace{.2cm}
    \centering
    \includegraphics[width=.9\columnwidth]{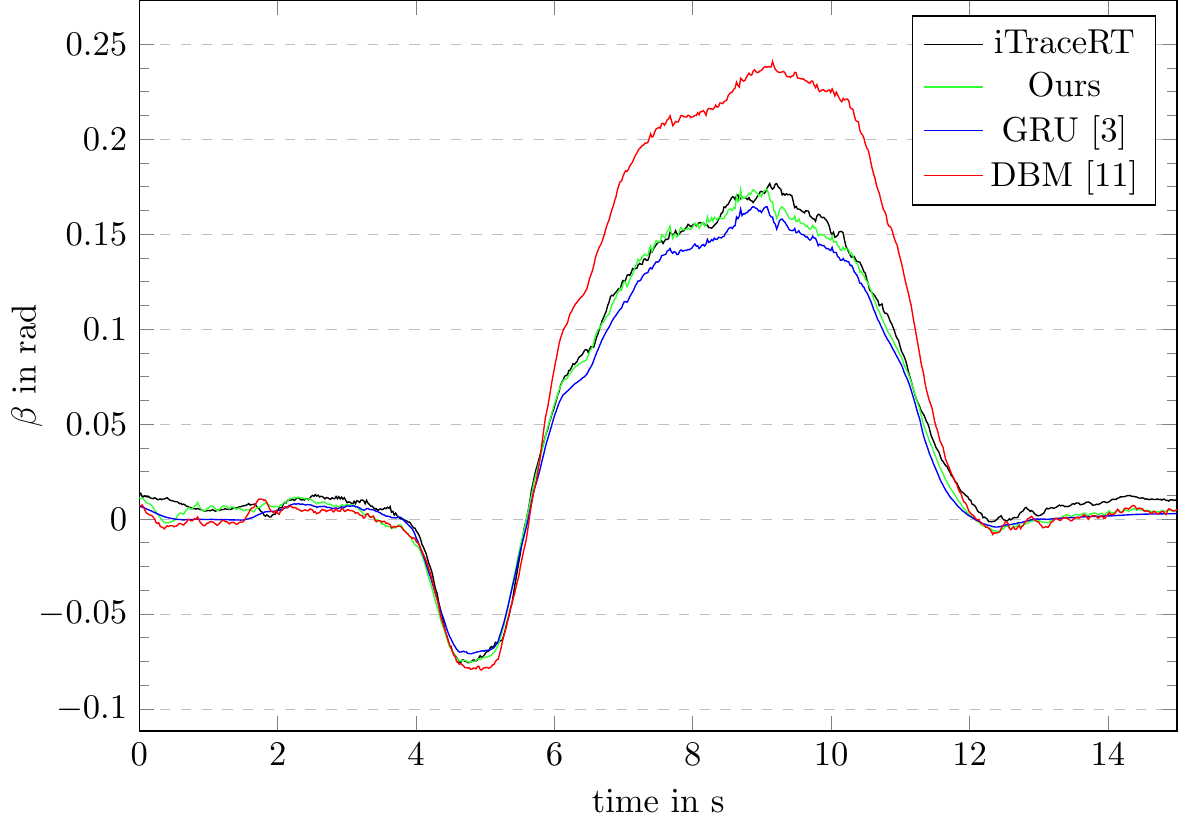}
    \caption{Side-slip angle estimation by the different observers for the high lateral acceleration point. The proposed approach has the closest estimations to the reference iTraceRT sensor.}
    \label{plot_81.fig}
\end{figure}

To evaluate the performance of the proposed observer with the variations of the  lateral acceleration, Fig. \ref{errors_81.fig} shows the errors of the observers in parallel to absolute values of the lateral acceleration. The plotted error is defined as:
\begin{equation}
    e^i_k = \left|\hat{\beta}^i_k-\beta^{\text{ref}}_k\right|
\end{equation}
with $k$ being the considered time step, $\hat{\beta}^i_k$ being the side-slip angle estimate by observer $i$ at time step $k$, and $\beta^{\text{ref}}_k$ being the reference side-slip angle measured by  the iTraceRT at time step $k$.
The plot shows that the proposed approach presents the lowest errors. The errors of all the observers increase with the rise of the lateral acceleration: the EKF based on the dynamic bicycle model \cite{reina_vehicle_2019} shows the highest sensitivity to high lateral accelerations, its errors grow up to $\SI{0.081}{rad}$; the hybrid GRU-based approach \cite{graber_hybrid_2019} presents a better performance but its errors can reach that of the EKF \cite{reina_vehicle_2019} (e.g. at $t=16 \si{\second}$);  the proposed approach shows robustness to high accelerations and maintains the lowest errors along the trajectory. 
A sample of the performance of the observers for the highest lateral acceleration point is shown in Fig. \ref{plot_81.fig} showing that the performance of the proposed method is the closest to the reference sensor. 

\begin{table}
 \vspace{.2cm}
\centering
\setlength\tabcolsep{16 pt}
\caption{MAE for the different observers calculated for the harsh maneuver ($a_y^{\text{max}} = 0.85g $). The errors of our approach are the lowest among the state-of-the-art observers. Check Table \ref{MAE1.tab} for the legend.\label{MAE3.tab}}
 % \resizebox{\columnwidth}{!}{
    \begin{tabular}{|c|c|c|c|c|c|}
      \hline
      State & DBM \cite{reina_vehicle_2019} &  GRU \cite{graber_hybrid_2019}  & Ours\\
      \hline
      $\beta$ \SI{}{(mrad)} & 9.79 & 7.48 & \textbf{4.42} \\
      \hline
      \end{tabular}
      % }
\end{table}

In brief, the proposed approach is able to deliver the most accurate estimations among the considered state-of-the-art approaches for both low dynamic and high dynamic maneuvers; it adapts to harsh maneuvers where other state-of-the-art approaches lose accuracy. 

\begin{figure}
    \centering
    \includegraphics[width=.9\columnwidth]{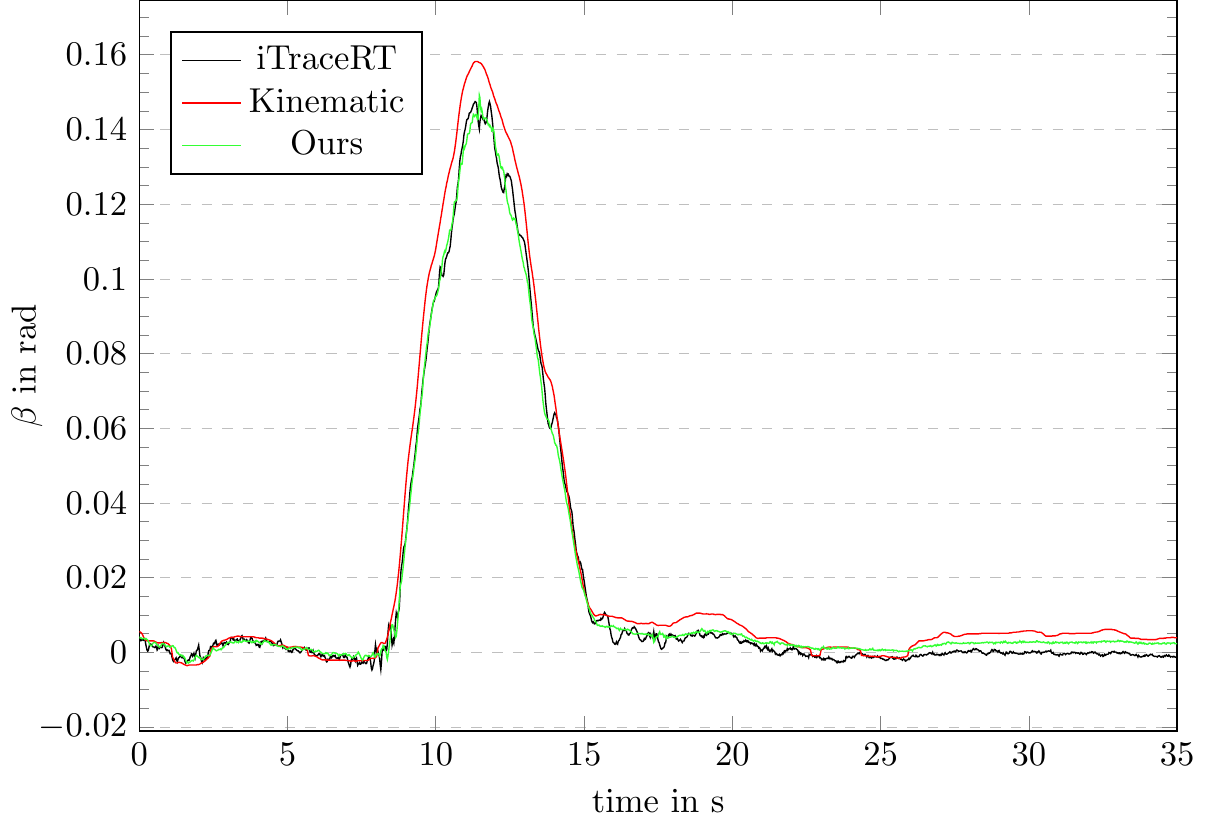}
    \caption{Comparison between the behavior of the kinematic side-slip angle, the proposed approach and the reference iTraceRT sensor: The proposed approach is able to reduce the offsets of the kinematic model and present results closer to the reference sensor.}
    \label{kinematicPlot2.fig}
\end{figure}

Having proved the accuracy of the proposed method, we compare the behavior of the kinematic bicycle side-slip angle to both the proposed approach estimations and the reference iTraceRT side-slip angle. A sample trajectory plot is shown in Fig.~\ref{kinematicPlot2.fig}. We see that our approach is able to correct the offsets available in the kinematic side-slip angle. This proves that the proposed hybrid observer takes advantage of the kinematic model but makes the necessary corrections to provide accurate estimations.

\section{Conclusion}\label{conclusion.sec}
In this paper, we presented a novel observer architecture for the estimation of the vehicle's side-slip angle. The proposed architecture takes as its inputs measurements from conventional car sensors in addition to a kinematic side-slip angle based on a physical model. We used a real vehicle to collect the needed data sets; the maneuvers were carefully performed to cover low acceleration and high acceleration scenarios. The presented approach was compared to state-of-the-art approaches for both low acceleration and near-limits maneuvers and was able to deliver the lowest errors for all considered scenarios. While the state-of-the-art methods lose accuracy with the increase of the harshness of the maneuver, the proposed approach adapts and delivers accurate estimations. 

Future work will investigate the advantages and limits of using a physical simple model along neural networks to observe the states of a vehicle; and the ability of neural networks to correct the inaccuracies of simplified vehicle models. 

\section*{Acknowledgement}
We would like to thank Prof. Markus Maurer at the Institute of Control Engineering, TU Braunschweig for enabling the collaboration that led to this joint paper. 

\bibliographystyle{ieeetr}
\bibliography{main}

\end{document}